%% file: main.tex
\def\year{2021}\relax
\documentclass[letterpaper]{article} 
\usepackage{aaai21}  
\usepackage{times}  
\usepackage{helvet} 
\usepackage{courier}  
\usepackage[hyphens]{url}  
\usepackage{graphicx} 
\usepackage{natbib}  
\usepackage{caption} 
\usepackage{mathrsfs,amsmath}
\usepackage{amssymb}
\usepackage{bm}
\usepackage{multirow}
\usepackage{colortbl,comment,xfrac}
\usepackage{amsthm}
\usepackage{multirow}
\usepackage{mathtools}

\input{defs}

\usepackage{paralist}
\usepackage[ruled,vlined]{algorithm2e}
\usepackage[caption = false]{subfig}

\newtheorem{definition}{Definition}
\newtheorem{lemma}{Lemma}

\title{Nystr\"{o}mformer: A Nystr\"{o}m-based Algorithm for Approximating Self-Attention}
\author {
    Yunyang Xiong \textsuperscript{\rm 1}$\quad$
    Zhanpeng Zeng \textsuperscript{\rm 1}$\quad$
    Rudrasis Chakraborty \textsuperscript{\rm 2}$\quad$ 
    Mingxing Tan \textsuperscript{\rm 3}\\
    Glenn Fung \textsuperscript{\rm 4}$\quad$
    Yin Li \textsuperscript{\rm 1}$\quad$
    Vikas Singh \textsuperscript{\rm 1}
    \\
}
\affiliations {
    \textsuperscript{\rm 1} University of Wisconsin-Madison $\quad$
    \textsuperscript{\rm 2} UC Berkeley $\quad$
     \textsuperscript{\rm 3} Google Brain $\quad$
    \textsuperscript{\rm 4} American Family Insurance\\
    yxiong43@wisc.edu, zzeng38@wisc.edu, rudra@berkeley.edu, tanmingxing@google.com, gfung@amfam.com, yin.li@wisc.edu, vsingh@biostat.wisc.edu
}
\begin{document}

\maketitle

\begin{abstract}
\input{abstract.tex}
\end{abstract}

\section{Introduction}
\input{intro.tex}

\section{Related Work}
\input{rel_work.tex}

\section{Nystr\"{o}m-Based Linear Transformers}
\input{nystrom_intro.tex}
\subsection{Nystr\"{o}m Method for Matrix Approximation} \label{sec:linear}
\input{nystrom_attention.tex}
\subsection{Linearized Self-Attention via Nystr\"{o}m  Method}
\label{sec:linearex}
\input{ex_nystrom_attention.tex}
\subsection{Analysis of Nystr\"{o}m Approximation}
\input{ex_nystrom_attention_ana.tex}
\subsection{Our Model: Nystr\"{o}mformer}
\input{arch.tex}

\section{Experiments}
\input{exp.tex}

\section{Conclusion}
\input{conclusion.tex}
\medskip

{\bf Acknowledgments.}
This work was supported in part by a American Family Insurance grant via American Family Insurance Data Science Institute at UW, NSF CAREER award RI 1252725 and UW CPCP (U54AI117924). We thank Denny Zhou, Hongkun Yu, and Adam Yu for valuable discussions. The paper also benefited from comments regarding typos and suggestions pointed out by Yannic Kilcher, Sebastian Bodenstein 
and Github user thomasw21. We thank Phil Wang and Lekton Zhang for making their implementation available at \text{\url{https://github.com/lucidrains/nystrom-attention}}.

\small
\bibliography{main}
\end{document}

%% file: defs.tex
\usepackage{stmaryrd}

\newcommand{\omitme}[1]{}

\newcommand{\Kc}{\mathcal{K}}

\newcommand{\Rb}{\mathbb{R}}

\newcommand{\Lleft}{\left(}
\newcommand{\Rright}{\right)}
\DeclarePairedDelimiterX{\norm}[1]{\lVert}{\rVert}{#1}

\def\Rb{\textbf{R}}



%% file: abstract.tex
Transformers have emerged as a powerful tool for a broad range of natural language processing tasks. A key component that drives the impressive performance of Transformers is the self-attention mechanism that encodes the influence or dependence of other tokens on each specific token. While beneficial, the quadratic complexity of self-attention on the input sequence length has limited its application to longer sequences -- a topic being actively studied in the community. To address this limitation, we propose Nystr\"{o}mformer -- a model that exhibits favorable scalability as a function of sequence length. Our idea is based on adapting the Nystr\"{o}m method to approximate  standard self-attention with $O(n)$ complexity. The scalability of Nystr\"{o}mformer enables 
application to longer sequences with thousands of tokens. We perform evaluations on multiple downstream tasks on the GLUE benchmark and IMDB reviews with standard sequence length, and find that our Nystr\"{o}mformer performs comparably, or in a few cases, even slightly better, than standard self-attention. On longer sequence tasks in the Long Range Arena (LRA) benchmark, Nystr\"{o}mformer performs favorably relative to other efficient self-attention methods. Our code is available at  \textit{\url{https://github.com/mlpen/Nystromformer}}.

%% file: intro.tex
Transformer-based models, such as BERT \cite{devlin2019bert} and GPT-3 \cite{brown2020language}, have been very successful in natural language processing (NLP), achieving state-of-the-art performance in machine translation \cite{vaswani2017attention}, natural language inference \cite{williams2018broad}, paraphrasing \cite{dolan2005automatically}, text classification \cite{howard2018universal}, question answering \cite{rajpurkar2016squad} and many other NLP tasks \cite{peters2018deep,radford2018improving}. A key feature of transformers is what is known as the self-attention mechanism \cite{vaswani2017attention}, where each token’s representation is computed from {\it all other} tokens. Self-attention enables interactions of token pairs across the full sequence and has been shown quite effective. 

Despite the foregoing advantages, self-attention also turns out to be a major efficiency bottleneck since it has a memory and time complexity of $O(n^2)$ where $n$ is the length of an input sequence. This leads to high memory and computational requirements for training large Transformer-based models. For example, training a BERT-large model \cite{devlin2019bert} will need 4 months using a single Tesla V100 GPU (equivalent to 4 days using a 4x4 TPU pod). Further, the $O(n^2)$ complexity makes it prohibitively expensive to train large Transformers with long sequences (e.g., $n=2048$). 

To address this challenge, several recent works have proposed strategies that avoid incurring the quadratic cost when dealing with longer input sequences. For example, \cite{dai2019transformer} suggests a trade-off between memory and computational efficiency. The ideas described in \cite{child2019generating,kitaev2019reformer} decrease the self-attention complexity to $O(n\sqrt{n})$ and $O(n\log n)$ respectively. In \cite{shen2018efficient,katharopoulos-et-al-2020,wang2020linformer}, self-attention complexity can be reduced to $O(n)$ with various approximation ideas, each with its 
own strengths and limitations. 

In this paper, we propose a $O(n)$ approximation, both in the sense of memory and time,  for self-attention. 
Our model, \textit{Nystr\"{o}mformer}, scales linearly with the input sequence length $n$. This is achieved by leveraging the celebrated Nystr\"{o}m method, repurposed for approximating self-attention. Specifically, our \textit{Nystr\"{o}mFormer} algorithm makes use of \textit{landmark} (or Nystr\"{o}m) points to reconstruct the softmax matrix in self-attention, thereby avoiding computing the $n\times n$ softmax matrix. We show that this yields a good approximation of the true self-attention.

To evaluate our method, we consider a transfer learning setting using Transformers, where models are first pretrained with a language modeling objective on a large corpus, and then finetuned on target tasks using supervised data \cite{devlin2019bert,liu2019roberta,lewis2019bart,wang2020linformer}. Following BERT \cite{devlin2019bert,liu2019roberta}, we pretrain our proposed model on English Wikipedia and BookCorpus \cite{zhu2015aligning} using a masked-language-modeling objective. We observe a similar performance to the baseline BERT model on English Wikipedia and BookCorpus. We then finetune our pretrained models on multiple downstream tasks in the GLUE benchmark \cite{wang2018glue} and IMDB reviews \cite{maas2011learning}, and compare our results to BERT in both accuracy and efficiency. Across all tasks, our model compares favorably to the vanilla pretrained BERT with significant speedups. 

Finally, we evaluate our model on tasks with longer sequence lengths from the Long Range Arena (LRA) benchmark \cite{tay2020long}. \textit{Nystr\"{o}mFormer} performs well compared to several recent efficient self-attention methods, including Reformer \cite{kitaev2019reformer}, Linformer \cite{wang2020linformer}, and Performer \cite{choromanski2020rethinking}, by margin of $\sim$3.4\% in average accuracy. We believe that the idea is a step towards resource efficient Transformers.

%% file: rel_work.tex
We briefly review relevant works on efficient Transformers, linearized Softmax kernels and Nystr\"{o}m-like methods.

\medskip
\noindent\textbf{Efficient Transformers}. Weight pruning \cite{michel2019sixteen}, weight factorization \cite{lan2019albert}, weight quantization \cite{zafrir2019q8bert} or knowledge distillation \cite{sanh2019distilbert} are several strategies that have been proposed 
to improve memory efficiency in Transformers. The use of a new pretraining objective in \cite{clark2019electra},  product-key attention in \cite{lample2019large}, and the Transformer-XL model in  \cite{dai2019transformer} have shown 
how the overall compute requirements can be reduced. In \cite{child2019generating}, a sparse factorization of the attention matrix was used for reducing the overall complexity from quadratic to $O(n\sqrt{n})$ for generative modeling of long sequences. In \cite{kitaev2019reformer}, the Reformer model further reduced the complexity to $O(n\log n)$ via  locality-sensitive-hashing (LSH). This relies on performing fewer dot product operations overall by 
assuming that the keys need to be identical to the queries. Recently, in \cite{wang2020linformer}, the Linformer model suggested the use of random projections based on the JL lemma to reduce the complexity to $O(n)$ with a linear projection step. The Longformer model in \cite{Beltagy2020Longformer} achieved a $O(n)$ complexity using a local windowed attention and a task-motivated global attention for longer documents, while BIGBIRD \cite{zaheer2020big} used a sparse attention mechanism. There are also other existing approaches to improve optimizer efficiency, such as micro-batching \cite{huang2019gpipe} and gradient checkpointing \cite{chen2016training}. Concurrently with our developments, the Performer model proposed in \cite{choromanski2020rethinking} made use of positive orthogonal random features to approximate softmax attention kernels with $O(n)$ complexity. 

\medskip
\noindent\textbf{Linearized Softmax}. In \cite{blanc2018adaptive}, an adaptive sampled softmax with a kernel based sampling was shown to speed up training. It involves sampling only some of the classes at each training step using a linear dot product approximation. In \cite{rawat2019sampled}, the Random Fourier Softmax (RF-softmax) idea uses random Fourier features to perform efficient sampling from an approximate softmax distribution for normalized embedding. In \cite{shen2018efficient,katharopoulos-et-al-2020}, linearizing the softmax attention in transformers was based on heuristically separating keys and queries in a linear dot product approximation. While the idea is interesting, the approximation error to the softmax matrix in self-attention can be large in some cases. 
The lambda layers in \cite{bello2021lambdanetworks}, can also be thought of as an efficient relative attention mechanism.

\medskip
\noindent\textbf{Nystr\"{o}m-like methods}. Nystr\"{o}m-like methods sample columns of the matrix to achieve a close approximation to the original matrix. The Nystr\"{o}m method \cite{baker1977numerical} was
developed as a way of discretizing an integral equation with a simple quadrature rule and remains  a widely used approach for approximating the kernel matrix with a given sampled subset of columns \cite{williams2001using}. Many variants such as Nystr\"{o}m with $k$-means \cite{zhang2008improved,zhang2010clustered}, randomized Nystr\"{o}m \cite{li2010making}, Nystr\"{o}m with spectral shift \cite{wang2014improving}, Nystr\"{o}m with pseudo landmarks, prototype method \cite{wang2013improving,wang2016towards}, fast-Nys \cite{si2016computationally}, and MEKA \cite{si2017memory}, ensemble Nystr\"{o}m \cite{kumar2009ensemble} have been proposed 
for specific improvements over the basic Nystr\"{o}m approximation. In \cite{nemtsov2016matrix}, the Nystr\"{o}m method was extended to deal with a general matrix (rather than a symmetric matrix). The authors in \cite{musco2017recursive} introduced the RLS-Nyström method, which proposes a recursive sampling approach to accelerate landmark points sampling. \cite{fanuel2019nystr} developed DAS (Deterministic Adaptive Sampling) and RAS (Randomized Adaptive Sampling) algorithms to promote diversity of landmarks selection. 

The most related ideas to our development are  \cite{wang2013improving,musco2017recursive}. These approaches are designed for general matrix approximation (which accurately reflects our setup) while only sampling a subset of columns and rows. However, directly applying these methods to approximate a softmax matrix used by self-attention does not directly reduce the computational complexity. This is because that even accessing a subset of columns or rows of a softmax matrix will require the calculation of all elements in the full matrix before the softmax function. And calculating these entries will incur a quadratic cost. Nonetheless, inspired by the key idea of using a subset of columns to reconstruct the full matrix, we propose a Nystr\"{o}m approximation with $O(n)$ complexity tailored for the softmax matrix, for approximating self-attention efficiently.

%% file: nystrom_intro.tex
In this section, we start by briefly reviewing self-attention, then discuss the basic idea of Nystr\"{o}m approximation method for the softmax matrix in self-attention, and finally adapting this idea to achieve our proposed construction.

\subsection{Self-Attention}
{\bf What is self-attention?} Self-attention calculates a weighted average of feature representations with the weight proportional to a similarity score between pairs of representations. Formally, an input sequence of $n$ tokens of dimensions $d$, $X \in \Rb^{n\times d}$, is projected using three matrices $W_Q \in \Rb^{d \times d_q}$, $W_K \in \Rb^{d \times d_k}$, and $W_V \in \Rb^{d\times d_v}$ to extract feature representations $Q$, $K$, and $V$, referred to as query, key, and value respectively with $d_k = d_q$. The outputs $Q$, $K$, $V$ are computed as 
\begin{equation}
\small
\begin{split}
    Q  = XW_Q, \quad
    K  = XW_K, \quad
    V  = XW_V.
\end{split}
\end{equation}
So, self-attention can be written as,
\begin{equation}\label{eq:trueatten}
\small
     D(Q, K, V) = SV = \text{softmax}\Lleft\frac{QK^T}{\sqrt{d_q}}\Rright V,
\end{equation}
where $\text{softmax}$ denotes a {\it row-wise} softmax normalization function. Thus, each element in the softmax matrix $S$ depends on all other elements in the same row.

\smallskip
\noindent {\bf Compute cost of self-attention}. The self-attention mechanism requires calculating $n^2$ similarity scores between each pair of tokens, leading to a complexity of $O(n^2)$ for both memory and time. Due to this quadratic dependence on the input length, the application of self-attention is limited to short sequences (e.g., $n<1000$).  This is a 
{\it key motivation} for a resource-efficient self-attention module.

%% file: nystrom_attention.tex
The starting point of our work is to reduce the computational cost of self-attention in Transformers using the Nystr\"{o}m method, widely adopted for matrix approximation \cite{williams2001using,drineas2005nystrom,wang2013improving}. Following \cite{wang2013improving}, we describe a potential strategy and its challenges for using the Nystr\"{o}m method to approximate the softmax matrix in self-attention by sampling a subset of columns and rows. 

Denote the softmax matrix used in self-attention $S = \text{softmax}\Lleft \frac{QK^T}{\sqrt{d_q}} \Rright \in \Rb^{n\times n}$. $S$ can be written as
\begin{equation}\label{eq:gendec}
\small
    S = \text{softmax}\Lleft \frac{QK^T}{\sqrt{d_q}} \Rright =
    \begin{bmatrix}
    A_S & B_S \\
    F_S & C_S
    \end{bmatrix},
\end{equation}
where $A_S \in \Rb^{m\times m}, B_S \in \Rb^{m\times (n - m)}, F_S \in \Rb^{(n - m) \times m}$ and $C_S \in \Rb^{(n-m)\times (n -m)}$. $A_S$ is designated to be our sample matrix by sampling $m$ columns and rows from $S$. 

\smallskip
\noindent \textbf{Quadrature technique.} $S$ can be approximated via the basic quadrature technique of the Nystr\"{o}m method. It begins with the singular value decomposition (SVD) of the sample matrix, $A_S = U\Lambda V^T$, where $U, V \in \Rb^{m \times m}$ are orthogonal matrices, $\Lambda \in \Rb^{m \times m}$ is a diagonal matrix. Based on the out-of-sample columns approximation \cite{wang2013improving}, the explicit Nystr\"{o}m form of $S$ can be reconstructed with $m$ columns and $m$ rows from $S$, 

\begin{align}\label{eq:gennys}
\small
    \hat{S} = 
    \begin{bmatrix}
    A_S & B_S \\
    F_S & F_SA_S^{+}B_S
    \end{bmatrix}
    =
    \begin{bmatrix}
    A_S \\
    F_S
    \end{bmatrix}
    A_S^{+}
        \begin{bmatrix}
    A_S \quad B_S
    \end{bmatrix},
\end{align}
where $A_S^{+}$ is the Moore-Penrose inverse of $A_S$. $C_S$ is approximated  by $F_SA_S^{+}B_S$. Here, \eqref{eq:gennys} suggests that the $n\times n$ matrix $S$ can be reconstructed by sampling $m$ rows ($A_S, B_S$) and $m$ columns ($A_S, F_S$) from $S$ and finding the Nystr\"{o}m approximation $\hat{S}$. 

\smallskip
\noindent \textbf{Nystr\"{o}m approximation for softmax matrix.} We briefly discuss how to construct the out-of-sample approximation for the softmax matrix in self-attention using the 
standard Nystr\"{o}m method. Given a query $q_i$ and key $k_j$, 
let 
\begin{align*}
\small
    \Kc_{K}(q_i) = \text{softmax}{\Lleft \frac{q_iK^T}{\sqrt{d_q}}\Rright}; ~~
\Kc_{Q}(k_j) = \text{softmax}{\Lleft\frac{Qk_j^T}{\sqrt{d_q}}\Rright}
\end{align*}
where $ \Kc_{K}(q_i) \in \Rb^{1 \times n}$ and $\Kc_{Q}(k_j) \in \Rb^{n \times 1}$. We can then construct
\begin{align*}
\small
\phi_{K}(q_i) & = \Lambda^{-\frac{1}{2}} V^T [\Kc^T_{K}(q_i)]_{m\times 1}  \\ \phi_{Q}(k_j) & = \Lambda ^{-\frac{1}{2}}U^T[\Kc_{Q}(k_j)]_{m\times 1}   
\end{align*}
where $[\cdot]_{m\times 1}$ refers to calculating the full $n \times 1$ vector and then taking the first $m \times 1$ entries. With $\phi_{K}(q_i)$ and $\phi_{Q}(k_j)$ available in hand, the entry of $\hat{S}$ for standard Nystr\"{o}m approximation is calculated as, 
\begin{align}\label{eq:standardsoft}
\small
    \hat{S}_{ij} = \phi_{K}(q_i)^T\phi_{Q}(k_j), \forall i = 1, \dots, n, j = 1, \dots, n
\end{align}
In matrix form, $\hat{S}$ can be represented as,
\begin{equation}\label{eq:softmaxmt}
\resizebox{.42\textwidth}{!} 
{
    $\hat{S} = \left[\text{softmax}\Lleft\frac{QK^T}{\sqrt{d_q}}\Rright\right]_{n\times m} A_S^{+}
    \left[\text{softmax}\Lleft\frac{QK^T}{\sqrt{d_q}}\Rright\right]_{m\times n}$
}
\end{equation}
where $\left[\cdot\right]_{n\times m}$ refers to  taking $m$ columns from $n \times n$ matrix and $\left[\cdot\right]_{m\times n}$ refers to taking $m$ rows from $n \times n$ matrix. This representation is the application of  \eqref{eq:gennys} for softmax matrix approximation in self-attention. $\begin{bmatrix} A_S \\ F_S \end{bmatrix}$ in \eqref{eq:gennys} corresponds to the first $n\times m$ matrix in  \eqref{eq:softmaxmt} and $\begin{bmatrix}
    A_S \quad B_S
    \end{bmatrix}$ in \eqref{eq:gennys} corresponds to the last $n\times m$ matrix in  \eqref{eq:softmaxmt}. 
More details of the matrix representation is available in the supplement.

\begin{figure}[t!]
\centering
\includegraphics[width=0.45\textwidth]{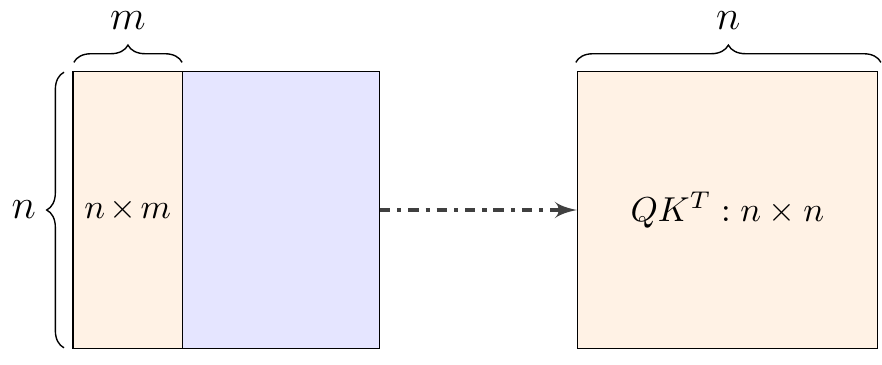}
\vspace{-7pt}
\caption{\label{fig:softmaxsubcol}\footnotesize A key challenge of Nystr\"{o}m approximation. The orange block on the left shows a $n \times m$ sub-matrix of $S$ used by Nystr\"{o}m matrix approximation in \eqref{eq:gennys}. Computing the sub-matrix, however, requires all entries in the $n\times n$ matrix before the softmax function ($QK^T$). Therefore, a direct application of Nystr\"{o}m approximation is problematic.}
\end{figure}

\smallskip
\noindent \textbf{A key challenge of Nystr\"{o}m approximation.} Unfortunately, \eqref{eq:gennys} and \eqref{eq:softmaxmt} require calculating all entries in $Q K^T$ due to the softmax function, even though the approximation only needs to access a subset of the columns of $S$, i.e., $\begin{bmatrix} A_S \\ F_S \end{bmatrix}$. The problem arises due to the denominator within the row-wise softmax function. Specifically, computing an element in $S$ requires a summation of the exponential of all elements in the same row of $Q K^T$. Thus, calculating $\begin{bmatrix} A_S \\ F_S \end{bmatrix}$ needs accessing the full $Q K^T$, shown in Fig. \ref{fig:softmaxsubcol},  and directly applying Nystr\"{o}m approximation as in  \eqref{eq:gennys} is not attractive.   

%% file: ex_nystrom_attention.tex
We now adapt the Nystr\"{o}m method to approximately calculate the full softmax matrix $S$.
The basic idea is to use landmarks $\Tilde{K}$ and $\Tilde{Q}$ from key $K$ and query $Q$ to derive an efficient Nystr\"{o}m approximation without accessing the full $Q K^T$. When the number of landmarks, $m$, is much smaller than the sequence length $n$, our Nystr\"{o}m approximation scales linearly w.r.t. input sequence length in the sense of both memory and time. %

Following the Nystr\"{o}m method, we also start with the SVD of a smaller matrix, $A_S$, and apply the basic quadrature technique. But {\em instead} of subsampling the matrix after the softmax operation -- 
as one should do in principle -- 
the main modification is to select landmarks $\Tilde{Q}$ from queries $Q$ and $\Tilde{K}$ from keys $K$ {\em before} softmax and then form a $m\times m$ matrix $A_S$ by applying the softmax operation on 
the landmarks. We also form the matrices corresponding to the left and right matrices 
in \eqref{eq:gennys} 
using landmarks $\Tilde{Q}$ and $\Tilde{K}$. 
This provides a $n\times m$ matrix and $m\times n$ matrix respectively.
With these three $n\times m, m \times m, m \times n$ matrices we constructed, our Nystr\"{o}m approximation of the $n\times n$ matrix $S$ involves the multiplication of three matrices as in \eqref{eq:gennys}. 

In the description that follows, we first define the matrix form of landmarks. Then, based on the landmarks matrix, we form the three matrices needed for our approximation. 

\begin{definition}
Let us assume that the selected landmarks for inputs 
$Q = [q_1; \dots; q_n]$ 
and $K = [k_1; \dots; k_n]$ are $\{\Tilde{q_j}\}_{j = 1}^m$ and $\{\Tilde{k_j}\}_{j = 1}^m$
respectively. 
We denote the matrix form of the corresponding landmarks as 
\begin{align*}
\small
&\text{For } \{\Tilde{q_j}\}_{j = 1}^m, \quad \Tilde{Q} = [\Tilde{q_1}; \dots; \Tilde{q_m}] \in \Rb^{m \times d_q}\\
&\text{For } \{\Tilde{k_j}\}_{j = 1}^m, \quad \Tilde{K} = [\Tilde{k_1}; \dots; \Tilde{k_m}] \in \Rb^{m \times d_q}    
\end{align*}
\end{definition} 

The corresponding $m\times m$ matrix is generated by 
\begin{equation*}
\small
A_S = \text{softmax}\Lleft\frac{\Tilde{Q}\Tilde{K}^T}{\sqrt{d_q}} \Rright 
\text{ where } A_S = U_{m \times m} \Lambda_{m \times m} V_{m \times m}^T
\end{equation*}
Note that in the SVD decomposition of $A_S$, 
$U_{m\times m}$ and $V_{m \times m}$ are orthogonal matrices. 

\begin{figure}[t]
\centering
\includegraphics[width=0.49\textwidth]{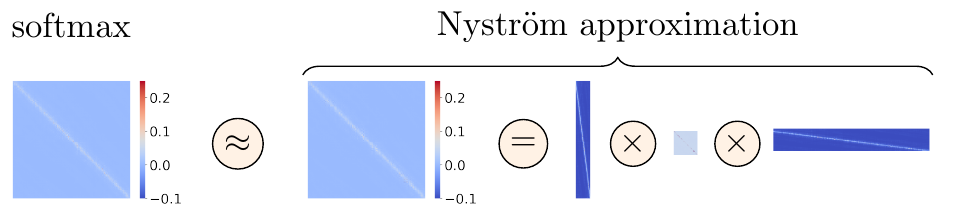}
\caption{\label{fig:nystromformer}\footnotesize Illustration of a Nystr\"{o}m approximation of softmax matrix in self-attention. The left image shows the true softmax matrix used in self-attention and the right images show its Nystr\"{o}m approximation. Our approximation is computed via multiplication of three matrices.}
\end{figure}

Similar to the out-of-sample approximation procedure for the standard Nystr\"{o}m scheme described above, given a query $q_i$ and key $k_j$, let 
\begin{align*}
\small
    \Kc_{\Tilde{K}}(q_i) = \text{softmax}{\Lleft \frac{q_i\Tilde{K}^T}{\sqrt{d_q}}\Rright}; ~~
\Kc_{\Tilde{Q}}(k_j) = \text{softmax}{\Lleft\frac{\Tilde{Q}k_j^T}{\sqrt{d_q}}\Rright},
\end{align*}
where $ \Kc_{\Tilde{K}}(q_i) \in \Rb^{1 \times m}$ and $\Kc_{\Tilde{Q}}(k_j) \in \Rb^{m \times 1}$. We can then construct, 
\begin{align*}
\small
\phi_{\Tilde{K}}(q_i) & = \Lambda^{-\frac{1}{2}}_{m\times m} V_{m \times m}^T \Kc^T_{\Tilde{K}}(q_i) \\ \phi_{\Tilde{Q}}(k_j) & = \Lambda_{m\times m}^{-\frac{1}{2}}U_{m\times m}^T\Kc_{\Tilde{Q}}(k_j) 
\end{align*}
So, the entry for $\hat{S}$ depends on landmark matrices $\Tilde{K}$ and $\Tilde{Q}$ and is calculated as,
\begin{align}
\small
    \hat{S}_{ij} = \phi_{\Tilde{K}}(q_i)^T\phi_{\Tilde{Q}}(k_j), \forall i = 1, \dots, n, j = 1, \dots, n,
\end{align}
To derive the explicit Nystr\"{o}m form, $\hat{S}$, of the softmax matrix with the three $n\times m$, $m\times m$, $m \times n$ matrices, we assume that $A_S$ is non-singular first to guarantee that the above expression to define $\phi_{\Tilde{K}}$ and $\phi_{\Tilde{Q}}$ is 
meaningful. We will shortly relax this assumption to achieve the general form as \eqref{eq:gennys}.

When $A_S$ is non-singular,  
\begin{align}\label{eq:shat}
\small
\hat{S}_{ij} &= \phi_{\Tilde{K}}(q_i)^T \phi_{\Tilde{Q}}(k_j)\\
    &= \Kc_{\Tilde{K}}(q_i)V_{m\times m}\Lambda_{m\times m}^{-1}U^T_{m \times m} \Kc_{\Tilde{Q}}(k_j).
\end{align}
Let $W_m  = V_{m\times m}\Lambda^{-1}_{m\times m}U_{m \times m}^T$. Recall that a SVD of $A_S$ is $U_{m \times m} \Lambda_{m\times m} V_{m \times m}^T$, and so, $W_mA_S = I_{m\times m}$. Therefore, 
\begin{align}\label{eq:sij}
\small
    \hat{S}_{ij} = \Kc_{\Tilde{K}}(q_i)A_S^{-1} \Kc_{\Tilde{Q}}(k_j)
\end{align}

Based on \eqref{eq:sij}, we can rewrite it to have a similar form as \eqref{eq:gennys} (i.e., not requiring that $A_S$ is non-singular) as
\begin{equation}\label{eq:exactnys}
    \hat{S}_{ij} = \Kc_{\Tilde{K}}(q_i)^TA_S^{+}\Kc_{\Tilde{Q}}(k_j),
\end{equation}
where $A_S^{+}$ is a Moore-Penrose pseudoinverse of $A_S$. So, 
\begin{align}
\small
\hat{S}_{ij} = \text{softmax}\Lleft\frac{q_i\Tilde{K}^T}{\sqrt{d_q}}\Rright A_S^{+}\text{softmax}\Lleft\frac{\Tilde{Q}k_j^T}{\sqrt{d_q}}\Rright,    
\end{align}
for $i, j = \{1, \dots, n\}$. The Nystr\"{o}m form of the softmax matrix, $S=\text{softmax}\Lleft\frac{QK^T}{\sqrt{d_q}}\Rright$ is thus approximated as 
\begin{equation}\label{eq:softapprox}
\resizebox{.475\textwidth}{!} 
{
    $\hat{S} = \text{softmax}\Lleft\frac{Q\Tilde{K}^T}{\sqrt{d_q}}\Rright\Lleft\text{softmax}\Lleft\frac{\Tilde{Q}\Tilde{K}^T}{\sqrt{d_q}}\Rright\Rright^{+}
    \text{softmax}\Lleft\frac{\Tilde{Q}K^T}{\sqrt{d_q}}\Rright$
}
\end{equation}

Note that we arrive at \eqref{eq:softapprox} 
via an out-of-sample approximation similar to  \eqref{eq:gennys}. The difference is that in \eqref{eq:softapprox}, the landmarks are selected before the softmax operation to generate the out-of-sample approximation. This 
is a compromise but 
avoids the need to compute the full softmax matrix $S$ for a Nystr\"{o}m approximation. 
Fig. \ref{fig:nystromformer} illustrates the proposed Nystr\"{o}m approximation and Alg. \ref{alg:nyssoftmax} summarizes our method.

We now describe 
\begin{inparaenum}[\bfseries (a)]
\item the calculation of the Moore-Penrose inverse and 
\item the selection of landmarks.
\end{inparaenum}

\begin{algorithm}[t]
\SetAlgoLined
\SetKwInput{KwInput}{Input}
\SetKwInput{KwOutput}{Output}
\KwInput{Query $Q$ and Key $K$.}
\KwOutput{Nystr\"{o}m approximation of softmax matrix.}
 Compute landmarks from input $Q$ and landmarks from input $K$, $\Tilde{Q}$ and $\Tilde{K}$ as the matrix form \;
 Compute $\Tilde{F} = \text{softmax}(\frac{Q\Tilde{K}^T}{\sqrt{d_q}})$, $\Tilde{B} = \text{softmax}(\frac{\Tilde{Q}K^T}{\sqrt{d_q}})$ \;
 Compute $\Tilde{A} = \text{softmax}(\frac{\Tilde{Q}\Tilde{K}^T}{\sqrt{d_q}})^{+}$\;
 \KwRet $\Tilde{F}\times\Tilde{A}\times\Tilde{B}$ \;
 \caption{Pipeline for Nystr\"{o}m approximation of softmax matrix in self-attention}
 \label{alg:nyssoftmax}
\end{algorithm}

\smallskip
\noindent\textbf{Moore-Penrose inverse computation}. Moore-Penrose pseudoinverse can be calculated by using singular value decomposition. However, SVD is not very efficient on GPUs. To accelerate the computation, we use an iterative method from \cite{razavi2014new} to approximate the Moore-Penrose inverse via efficient matrix-matrix multiplications.
\begin{lemma}\label{lem:pseudoinverse}
For $A_S \in \Rb^{m \times m}$, the sequence $\{Z_j\}_{j = 0}^{j = \infty}$ 
generated by \cite{razavi2014new},
\begin{equation}\label{eq:approxmp}
\small
    Z_{j + 1} = \frac{1}{4}Z_j(13I - A_S Z_j(15I - A_S Z_j(7I - A_S Z_j)))
\end{equation}
converges to the Moore-Penrose inverse $A_S^{+}$ in the third-order with initial approximation $Z_0$ satisfying $||A_SA_S^+ - A_S Z_0|| < 1$.
\end{lemma}
We select $Z_0$ by $Z_0 = \sfrac{A_S^T}{(||A_S||_1||A_S||_{\infty})}$ where 
\begin{align*}
\small
    ||A_S||_1 = \max_j\sum_{i = 1}^{m}|(A_S)_{ij}|; ~~
    ||A_S||_{\infty} = \max_{i}\sum_{j = 1}^{n}|(A_S)_{ij}|,
\end{align*}
based on \cite{pan1991improved}.
This choice ensures that $||I - A_S Z_0||_2 < 1$. 
When $A_S$ is non-singular, $$||A_SA_S^+ - A_S Z_0||_2 = ||I - A_SZ_0||_2 < 1.$$ 
Without the non-singular constraint, the choice of initializing $Z_0$ provides a good approximation in our experiments. For all our experiments, we need to run about $6$ iterations in order to achieve a good approximation of the pseudoinverse. 

Let $A_S^{+}$ be approximated by $Z^{\star}$ with  \eqref{eq:approxmp}. Our Nystr\"{o}m approximation of $S$ can be written as
\begin{equation}\label{eq:approxsoftmax}
\small
    \hat{S} = \text{softmax}\Lleft\frac{Q\Tilde{K}^T}{\sqrt{d_q}}\Rright Z^{\star}\text{softmax}\Lleft\frac{\Tilde{Q}K^T}{\sqrt{d_q}}\Rright.
\end{equation}
Here, \eqref{eq:approxsoftmax} only needs matrix-matrix multiplications, thus the gradient computation is straight-forward.

\smallskip
\noindent\textbf{Landmarks selection}. Landmark points (inducing points \cite{lee2019set}) can be selected by using K-means clustering \cite{zhang2008improved,vyas2020fast}. However, the EM style of updates in K-means is less 
desirable during mini-batch training. 
We propose to simply use Segment-means similar to the local average pooling previously used in the NLP literature \cite{shen2018baseline}. Specifically, for input queries $Q = [q_1; \dots; q_n]$, we separate the $n$ queries into $m$ segments. As we can pad inputs to a length divisible to $m$, we assume $n$ is divisible by $m$ for simplicity. Let $l = \sfrac{n}{m}$, landmark points for $Q$ are calculated as shown in \eqref{eq:landmarks-keys}.
Similarly, for input keys $K=[k_1; \dots; k_n]$, landmarks are computed as shown below in \eqref{eq:landmarks-keys}.
\begin{align}\label{eq:landmarks-keys}
\small
\Tilde{q}_j = \sum_{i = (j - 1)\times l + 1}^{(j - 1) \times l + m} \frac{q_i}{m}, \quad
\Tilde{k}_j = \sum_{i = (j - 1)\times l + 1}^{(j - 1) \times l + m}\frac{k_i}{m}, 
\end{align}
where $j=1, \cdots, m$.
Segment-means requires a single scan of the sequence to compute the landmarks leading to a complexity of $O(n)$. We find that using $64$ landmarks is often sufficient to ensure a good approximation, although this  
depends on the application. 
More details regarding the landmark selection is provided in the supplement. 

\begin{figure}[t]
\centering
\begin{minipage}[c]{0.8\linewidth}
\centering True self-attention
\includegraphics[width=1.0\textwidth]{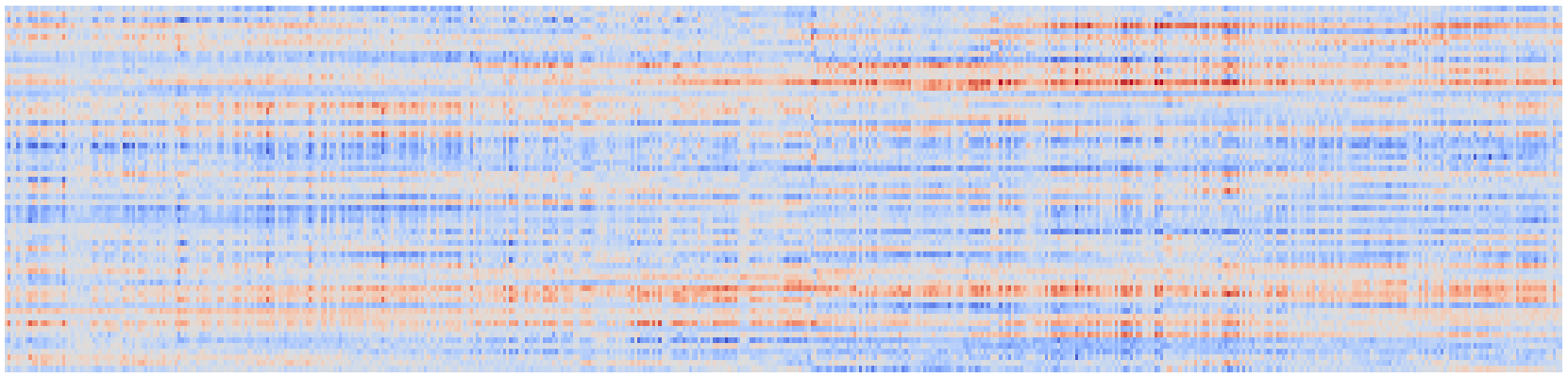}
\centering Nystr\"{o}m approximate self-attention
\includegraphics[width=1.0\textwidth]{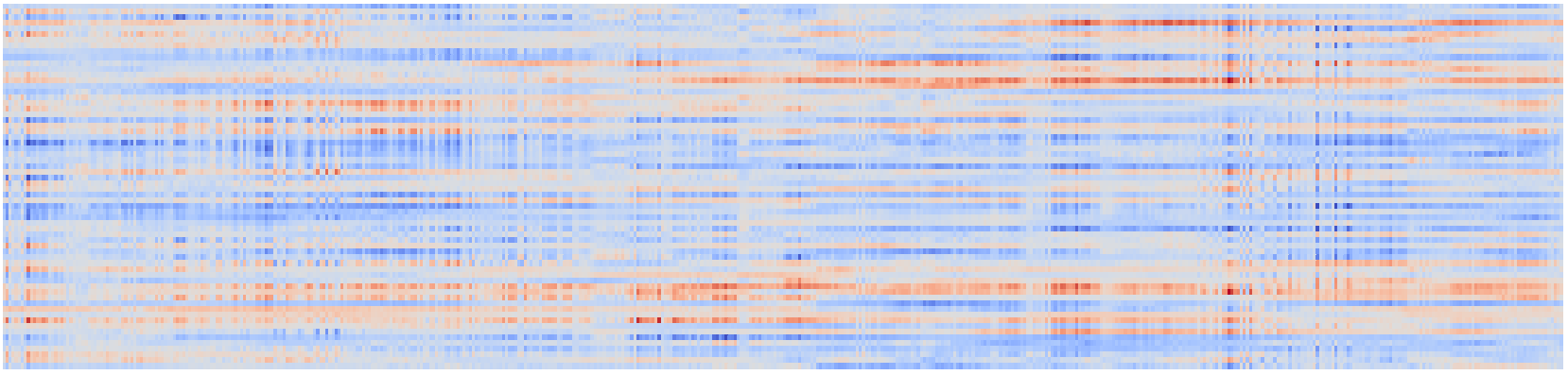}
\end{minipage}
\hfill
\begin{minipage}[c]{0.12\linewidth}
\centering 
\includegraphics[width=0.9\textwidth]{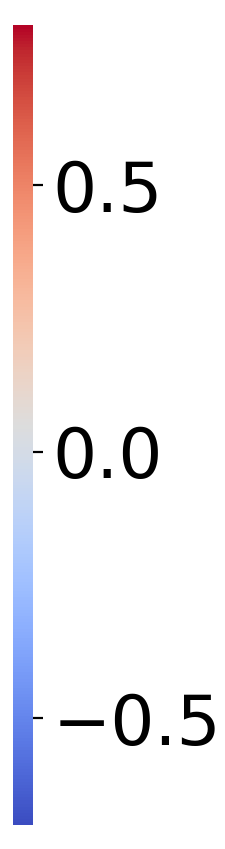}
\end{minipage}
\centering
\caption{\footnotesize\label{fig:nystrom_approximate}\footnotesize An example of Nystr\"{o}m approximation vs.\ ground-truth self-attention. Top: standard self-attention computed by \eqref{eq:trueatten}. Bottom: self-attention from our proposed Nystr\"{o}m approximation in \eqref{eq:approxatten}. We see that the attention patterns are quite similar.}
\end{figure}

\noindent\textbf{Approximate self-attention}.  With landmark points and pseudoinverse computed, the Nystr\"{o}m approximation of the softmax matrix can be calculated. By plugging in the Nystr\"{o}m approximation, we obtain a linearized version $\hat{S}V$, to approximate the true self-attention $SV$,
\begin{equation}\label{eq:approxatten}
\small
    \hat{S}V = \text{softmax}\Lleft\frac{Q\Tilde{K}^T}{\sqrt{d_q}}\Rright Z^{\star}\text{softmax}\Lleft\frac{\Tilde{Q}K^T}{\sqrt{d_q}}\Rright V.
\end{equation}
Fig. \ref{fig:nystrom_approximate} presents an example of the fidelity 
between Nystr\"{o}m approximate self-attention versus \ true self-attention.

\smallskip
\noindent\textbf{Complexity analysis}. 
We now provide a complexity analysis of the Nystr\"{o}m approximation, which needs to account for landmark selection, pseudoinverse calculation, and the matrix multiplications. Landmark selection using Segment-means takes $O(n)$. Iterative approximation of the pseudoinverse takes $O(m^3)$ in the worst case. The matrix multiplication first calculates $\text{softmax}(Q\Tilde{K}^T/\sqrt{d_q})\times Z^{\star}$ and $\text{softmax}(\Tilde{Q}K^T/\sqrt{d_q})\times V$, and then calculates the product $(\text{softmax}(Q\Tilde{K}^T/\sqrt{d_q})\times Z^{\star}) \times (\text{softmax}(\Tilde{Q}K^T/\sqrt{d_q})\times V)$. This costs $O(nm^2 + mnd_v + m^3 + nmd_v)$. The overall time complexity is thus $O(n + m^3 + nm^2 + mnd_v + m^3 + nmd_v)$. In terms of memory, storing the landmarks matrix $\Tilde{Q}$ and $\Tilde{K}$ involves a $O(md_q)$ cost and storing four Nystr\"{o}m approximation matrices has a $O(nm + m^2 + mn + nd_v)$ cost. Thus, the memory footprint is $O(md_q + nm + m^2 + mn + nd_v)$. When the number of landmarks $m \ll n$, the time and memory complexity of our Nystr\"{o}m approximation is $O(n)$, i.e., scales linearly w.r.t. the input sequence length $n$.

%% file: ex_nystrom_attention_ana.tex
\begin{figure*}[!t]
\centering
\includegraphics[width=0.75\textwidth]{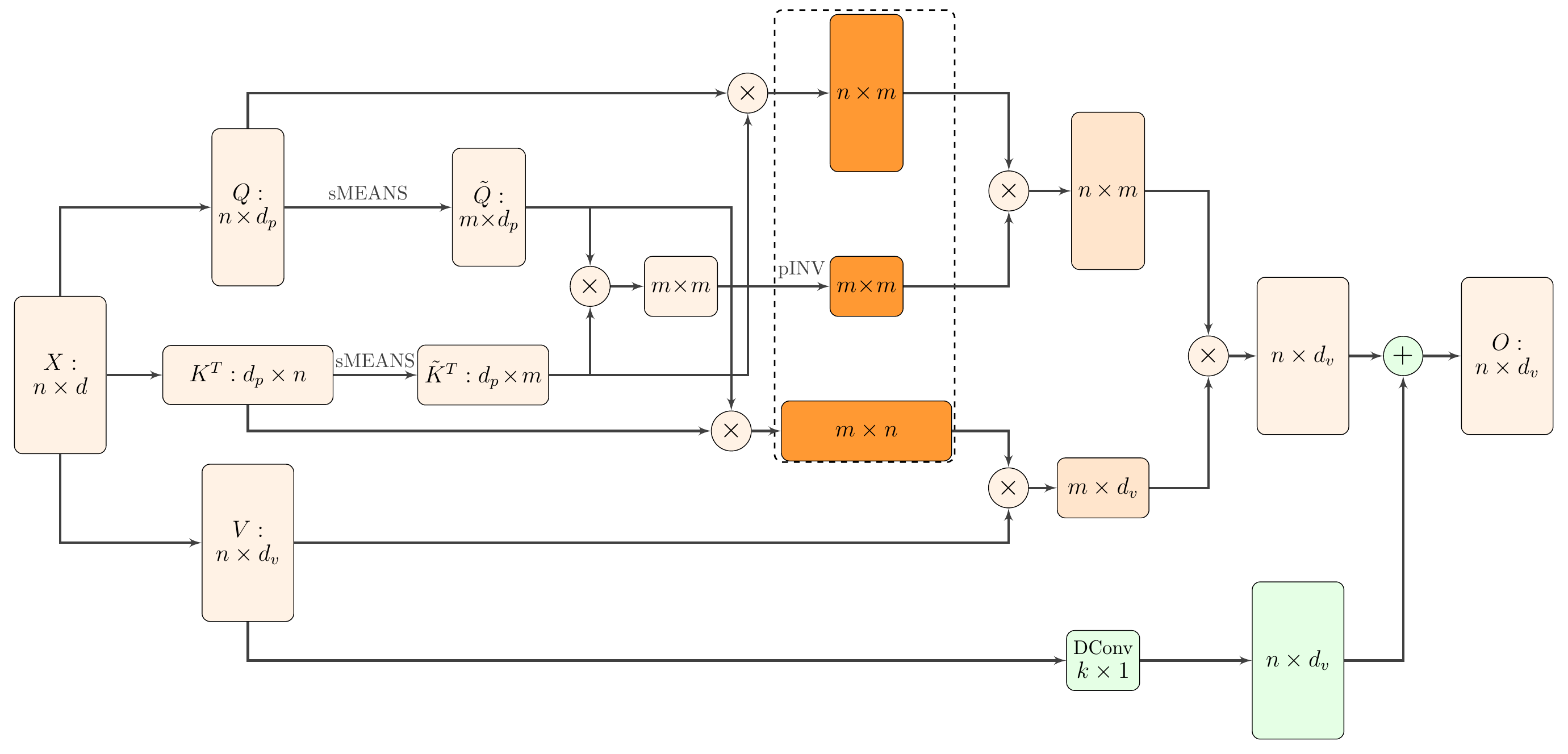}
\vspace*{-1.0em}
\caption{\label{fig:nystromtransformer}\footnotesize The proposed architecture of efficient self-attention via Nystr\"{o}m approximation. Each box represents an input, output, or intermediate
matrix. The variable name and the size of the matrix are inside box. $\times$ denotes matrix multiplication, and $+$ denotes matrix addition. The orange colored boxes are those matrices used in the Nystr\"{o}m approximation. The green boxes are the skip connection added in parrallel to the approximation.  The dashed bounding box illustrates the three matrices of Nystro\"{o}m approximate softmax matrix in self-attention in Eq. \ref{eq:approxsoftmax}. sMEANS is the landmark selection using Segment-means (averaging $m$ segments of input sequence). pINV is the iterative Moore-Penrose pseudoinverse approximation. And DConv denotes depthwise convolution.}
\end{figure*}

The following simple result analyzes an idealized setting and states that the Galerkin discretization of $\phi_{\Tilde{K}}(q)^T\phi_{\Tilde{Q}}(k)$ with the same set of landmark points, induces the same Nystr\"{o}m matrix, in particular, the same $n\times n$ Nystr\"{o}m approximation $\hat{S}_{ij}$. This result agrees with the discussion in \cite{bremer2012nystrom}. 
\begin{lemma}\label{lem:nlandmarks}
Given the input data set $Q = \{q_i\}_{i = 1}^n$ and $K = \{k_i\}_{i = 1}^n$, and the corresponding landmark point set $\Tilde{Q} = \{\Tilde{q_j}\}_{j = 1}^m$ and $\Tilde{K_j} = \{\Tilde{k}\}_{j = 1}^m$. 
Using \eqref{eq:approxatten}, the Nystr\"{o}m approximate self-attention converges to true self-attention if there exist landmarks points $\Tilde{q}_p$ and $\Tilde{k}_t$ such that $ \Tilde{q}_p = q_i$ and $\Tilde{k}_t = k_j$, $\forall i = 1, \dots, n, j = 1, \dots, n$. 
\end{lemma}

Lemma \ref{lem:nlandmarks} suggests that if the landmark points overlap sufficiently with the original data points, 
the approximation to self-attention will be good. 
While the condition here is problem dependent, 
we note that it is feasible to achieve an accurate approximation without using a large number of landmarks. This is because \cite{oglic2017nystrom} points out that the error of Nystr\"{o}m approximation depends on the spectrum of the matrix to be approximated and it decreases with the rank of the matrix. When this result is compared with the observation in \cite{wang2020linformer} where the authors suggest that self-attention is low-rank, stronger 
guarantees based on structural properties of the 
matrix that we wish to approximate are possible. 

%% file: arch.tex
\noindent\textbf{Architecture}. Our proposed architecture is shown in Fig.\ \ref{fig:nystromtransformer}. Given the input key $K$ and query $Q$, our model first uses Segment-means to compute landmark points as matrices $\Tilde{K}$ and $\Tilde{Q}$. With the landmark points, our model then calculates the Nystr\"{o}m approximation using approximate Moore-Penrose pseudoinverse. A skip connection of value $V$, implemented using a 1D depthwise convolution, is also added to the model to help the training.

%% file: exp.tex
We now present our experiments and results. Our experiments follow a transfer learning setting that consists of two stages. In the first stage, we train Nystr\"{o}mformer on a large-scale text corpus, and report the language modeling performance of our model on a hold-out validation set. In the second stage, we fine-tune the pre-trained Nystr\"{o}mformer across several different NLP tasks in GLUE benchmarks \cite{wang2019glue} and IMDB reviews \cite{maas2011learning}, and report the performance on individual dataset for each task. In both stages, we compare our results to a baseline Transformer model (BERT). In addition to language modeling, we also conduct experiments on long range context tasks in the Long Range Arena (LRA) benchmark.

\subsection{(Pre-)training of Language Modeling}

Our first experiment evaluates if our model can achieve similar performance with reduced complexity compared to a standard Transformer on language modeling. We introduce the dataset and evaluation protocol, describe implementation details, and finally present the results of our model.

\begin{table*}[!bt]
\centering
\resizebox{0.95\linewidth}{!}{
\begin{tabular}{c|c|c|c|c|c|c|c|c|c|c}
\hline
\multirow{3}{*}{self-attention} & \multicolumn{10}{c}{input sequence length n}                                                                                            \\ \cline{2-11} 
                                & \multicolumn{2}{c|}{512} & \multicolumn{2}{c|}{1024} & \multicolumn{2}{c|}{2048} & \multicolumn{2}{c|}{4096} & \multicolumn{2}{c}{8192} \\ \cline{2-11} 
                                & memory (MB)  & time(ms)  & memory (MB)  & time (ms)  & memory (MB)  & time (ms)  & memory (MB)  & time (ms)  & memory (MB)  & time (ms)  \\ \hline\hline
                Transformer                &   54 (1$\times$)        &    0.8 (1$\times $)       &     186 (1$\times$)      &     2.4 (1$\times$)      &   685 (1$\times$)        &  10.0 (1$\times$)      &    2620 (1$\times$)       &  32.9 (1$\times$)         &    10233 (1$\times$)       &    155.4 (1$\times$)       \\ \hline
                Linformer-256                &    41 (1.3$\times$)       &    0.7 (1.1$\times$)      &   81 (2.3$\times$)        &    1.3 (1.8$\times$)      &    165 (4.2$\times$)       &   2.7  (3.6$\times$)       &   366 (7.2$\times$)          &   5.3 (6.2$\times$)     &   635 (16.1$\times$)         &    11.3  (13.8$\times$)       \\ \hline
                Longformer-257 &    32.2 (1.7$\times$) & 2.4 (0.3$\times$)   &  65 (2.9$\times$)  & 4.6 (0.5$\times$)   & 130 (5.3$\times$)  & 9.2 (1.0$\times$)  & 263 (10.0$\times$) & 18.5 (1.8$\times$) & 455 (22.5$\times$) & 36.2 (4.3$\times$)\\ \hline \hline
                Nystr\"{o}mformer-64                &     35 (1.5$\times$)      &   0.7 (1.1$\times$)        &     63 (3.0$\times$)      &    1.3 (1.8$\times$)     &    118 (5.8$\times$)        &    2.7 (3.6$\times$)       &     229 (11.5$\times$)       &   5.9  (5.6$\times$)      &      450 (22.8$\times$)       &    12.3 (12.7$\times$)    \\ \hline
                Nystr\"{o}mformer-32                &     26 (2.1$\times$)      &   0.6 (1.2$\times$)        &     49 (3.8$\times$)      &    1.2 (1.9$\times$)     &    96 (7.1$\times$)        &    2.6 (3.7$\times$)       &     193 (13.6$\times$)       &   5.6  (5.9$\times$)      &      383 (26.7$\times$)       &    11.5 (13.4$\times$)    \\ \hline                
\end{tabular}}
\vspace*{-0.5em}
\caption{\footnotesize\label{tab:timememory} Memory consumption and running time results on various input sequence length. We report the average memory consumption (MB) and running time (ms) for one input instance with different input length through self-attention module. Nystr\"{o}mformer-64 denotes Nystr\"{o}mformer self-attention module using 64 landmarks and Nystr\"{o}mformer-32 denotes Nystr\"{o}mformer module using 32 landmarks. Linformer-256 denotes Linformer self-attention module using linear projection dimension $256$. Longformer-257 denotes Longformer self-attention using sliding window size $257 (128\times 2 + 1)$. Our Nystr\"{o}m self-attention offers favorable memory and time efficiency over standard self-attention and Longformer self-attention. With a length of $8192$, our model offers 1.2$\times$ memory saving and 3$\times$ speed-up over Longformer, and 1.7$\times$ memory saving over Linformer with similar running time. }
\end{table*}

\smallskip
\noindent \textbf{Dataset and metrics}. We consider BookCorpus plus English Wikipedia as the training corpus, which is further split into training (80\%) and validation (20\%) sets. Our model is trained using the training set. We report the masked-language-modeling (MLM) and sentence-order-prediction (SOP) accuracy on the validation set, and compare the efficiency (runtime/memory) to a baseline. 

\smallskip
\noindent \textbf{Baselines}. 
Our baseline is the well-known Transformer based model -- BERT \cite{devlin2019bert}. Specifically, we consider two variants of BERT:
\begin{itemize}
    \item \textbf{BERT-small} is a light weighted BERT model with 4 layers. We use BERT-small to compare to linear Transformers, including ELU linearized self-attention \cite{katharopoulos-et-al-2020} and Linformer \cite{wang2020linformer}. 
    \item \textbf{BERT-base} is the base model from \cite{devlin2019bert}. We use this model as our baseline when fine-tuning on downstream NLP tasks. 
\end{itemize}

Our Nystr\"{o}mformer replaces the self-attention in BERT-small and BERT-base using the proposed Nystr\"{o}m approximation. We acknowledge that several very recent articles \cite{zaheer2020big,Beltagy2020Longformer}, 
concurrent with our work, have also proposed efficient $O(n)$ self-attention for Transformers. An exhaustive comparison to a rapidly 
growing set of algorithms is prohibitive unless
extensive compute resources are freely available. 
Thus, we only compare runtime performance and the memory consumption of our method to Linformer \cite{wang2020linformer} and Longformer \cite{Beltagy2020Longformer} in Table~\ref{tab:timememory}. 

\smallskip
\noindent \textbf{Implementation details}. Our model is pre-trained with the masked-language-modeling (MLM) and sentence-order-prediction (SOP) objectives \cite{lan2019albert}. We use a batch size of 256, Adam optimizer with learning rate 1e-4, $\beta_1 = 0.9$, $\beta_2 = 0.999$, L2 weight decay of $0.01$, learning rate warm-up over the first 10,000 steps, and linear learning rate decay to update our model. Training BERT-base with $1$M update steps takes more than one week on 8 V100 GPUs. To keep 
compute costs reasonable, our baseline (BERT-base) and our model are trained with 0.5M steps. We also train our model with $\sim$0.25M steps, initialized from pre-trained BERT-base for speed-up. For BERT-small, we train for $0.1$M steps. More details are in the supplement. 

\begin{figure}[hbt!]
\vspace*{-0.5em}
\centering
\includegraphics[width=0.49\textwidth]{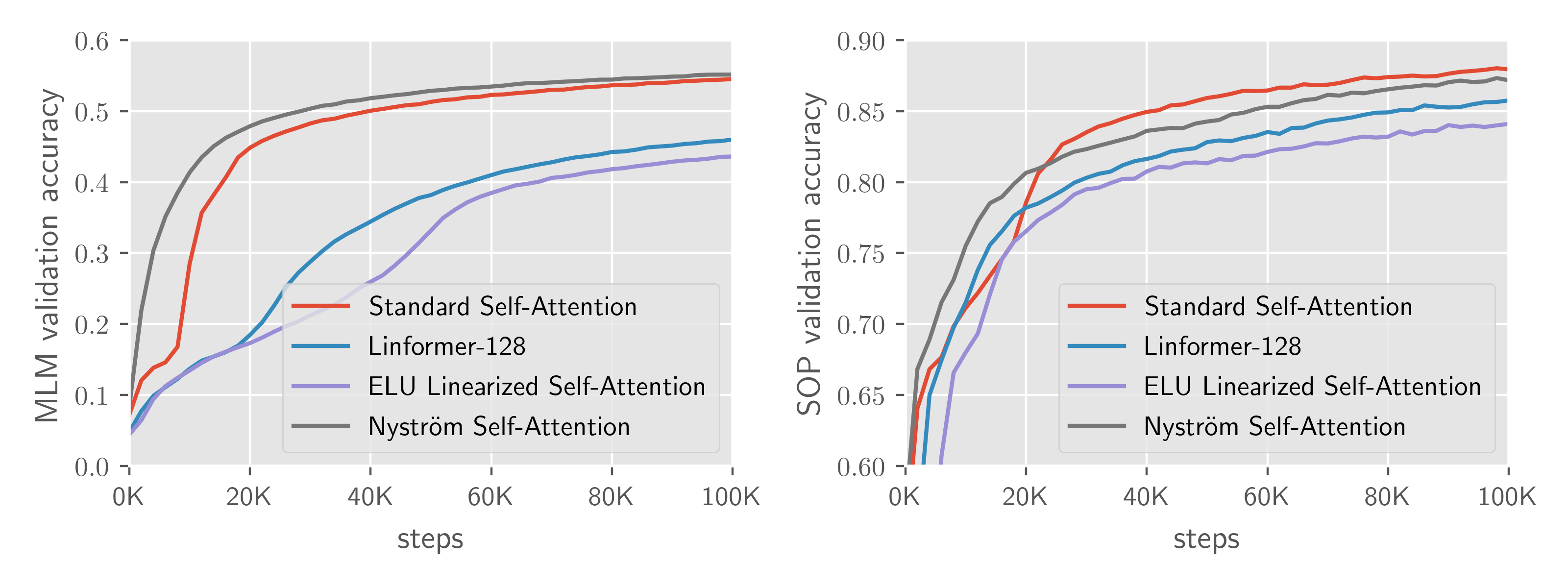}
\vskip -0.2in
\caption{\label{fig:validation-small}\footnotesize Results on masked-language-modeling (MLM) and sentence-order-prediction (SOP). On BERT-small, our Nystr\"{o}m self-attention is competitive to standard self-attention, outperforming Linformer and other linear self-attentions. }
\end{figure}

\begin{figure}[hbt!]
\vspace*{-0.5em}
\centering
\includegraphics[width=0.49\textwidth]{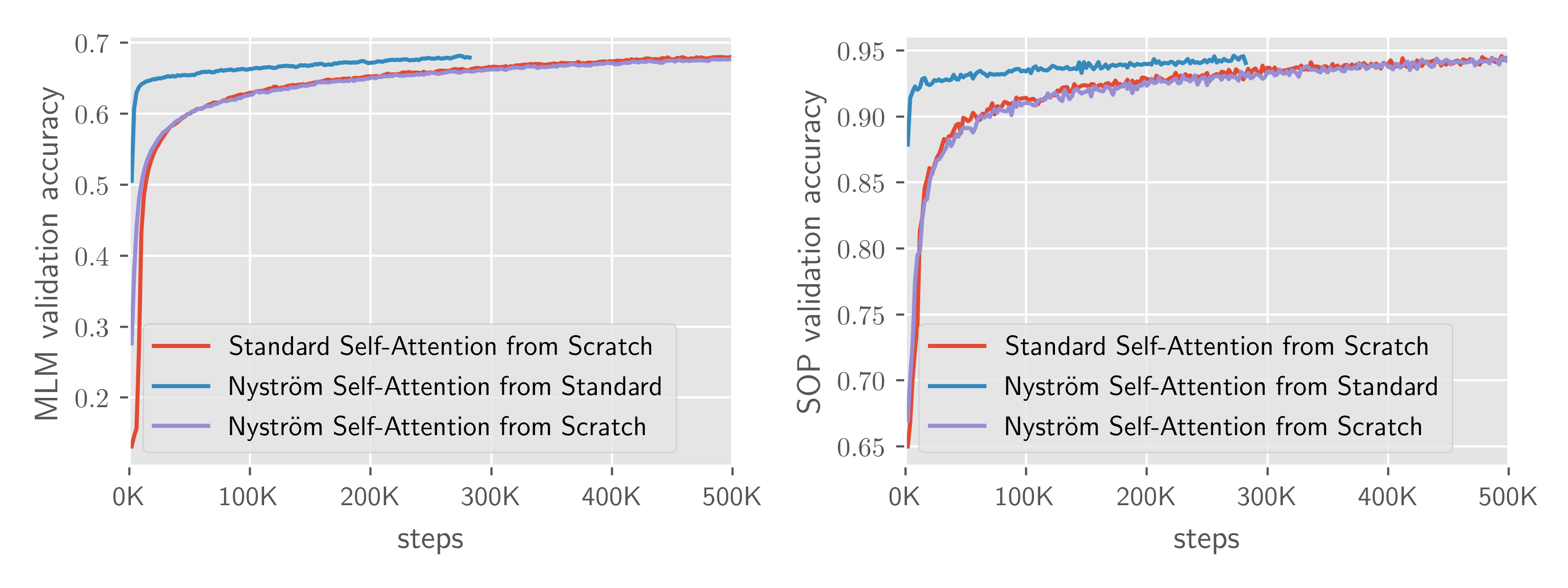}
\vskip -0.2in
\caption{\label{fig:validation}\footnotesize Results on MLM and SOP. We report MLM and SOP validation accuracy for each training step. BERT-base (from scratch) is trained with 0.5M steps, our Nystr\"{o}m (from scratch) is trained with 0.5M steps as BERT-base (from scratch), and our Nystr\"{o}mformer (from standard) is trained with $\sim$0.25M steps initialized from pretrained BERT-base. Our Nystr\"{o}m self-attention is competitive with standard self-attention, BERT-base, and initializing from pretrained BERT-base helps speed up training.}
\end{figure}

\smallskip
\noindent \textbf{Results on accuracy and efficiency}. We report the validation accuracy and inference efficiency of our model and compare the results to transformer based models. In Fig. \ref{fig:validation-small} and \ref{fig:validation}, we plot MLM and SOP pre-training validation accuracy, which shows that Nystr\"{o}former is comparable to a standard transformer and outperforms other variants of efficient transformers. We also note the computation and memory efficiency of our model in Table \ref{tab:timememory}. To evaluate the inference time and memory efficiency, we generate random inputs for self-attention module with sequence length $n \in [512, 1024, 2048, 4096, 8192]$. All models are evaluated on the same machine setting with a Nvidia 1080Ti and we report the improved inference speed and memory savings. 

\begin{table}[!tbh]
\small
\centering
\resizebox{0.99\linewidth}{!}{
\begin{tabular}{c|c|c|c|c|c|c}
\hline
Model          & SST-2 & MRPC & QNLI & QQP & MNLI m/mm & IMDB \\ \hline\hline
BERT-base      & 90.0  & 88.4 & 90.3 & 87.3 &  82.4/82.4 &  93.3 \\
\hline\hline
Nystr\"{o}mformer & 91.4 & 88.1 & 88.7 & 86.3 & 80.9/82.2 &  93.2 \\
\end{tabular}}
\hspace{0.1em}
\vspace*{-0.5em}
\caption{\footnotesize\label{tab:devset} Results on natural language understanding tasks. We report F1 score for MRPC and QQP and accuracy for others. Our Nystr\"{o}mformer performs competitively with BERT-base. }
\end{table}
\begin{table*}[!bth]
\small
\centering
\resizebox{0.85\linewidth}{!}{
\begin{tabular}{c|c|c|c|c|c|c}
\hline
Model          & ListOps (2K) & Text (4K) & Retrieval (4K) & Image (1K) &  Pathfinder (1K) & Avg \\ \hline\hline
Standard \cite{vaswani2017attention} & 37.10 & 65.02 & 79.35 & 38.20 & 74.16 & 58.77 \\
\hline\hline
Reformer \cite{kitaev2019reformer} & 19.05 & 64.88 & 78.64 & 43.29 & 69.36 & 55.04 \\
Linformer \cite{wang2020linformer} & 37.25 & 55.91 & 79.37 & 37.84 & 67.60 & 55.59 \\
Performer \cite{choromanski2020rethinking} & 18.80 & 63.81 & 78.62 & 37.07 & 69.87 & 53.63 \\
\hline\hline
Nystr\"{o}mformer (ours) & 37.15 & 65.52 & 79.56 & 41.58 & 70.94 & \textbf{58.95} \\
\end{tabular}}
\hspace{0.1em}
\vspace*{-0.5em}
\caption{\footnotesize\label{tab:lraset} Results on Long Range Arena (LRA) benchmark using our PyTorch implementation. We report classification accuracy for each individual task and average accuracy across all tasks. Our Nystr\"{o}mformer performs competitively with standard self-attention, and significantly outperforms Reformer, Linformer, and Performer. While we achieve consistent results reported in \cite{tay2020long} for most tasks in our PyTorch reimplementation, the performance on Retrieval task is higher for all models following the hyperparameters in \cite{tay2020long}.}
\end{table*}
\subsection{Fine-tuning on Downstream NLP tasks}
Our second experiment is designed to test the generalization ability of our model on downstream NLP tasks. To this end, we fine-tune the pretrained model across several NLP tasks.

\smallskip
\noindent \textbf{Datasets and metrics}. We consider the datasets of SST-2 \cite{socher2013recursive}, MRPC \cite{dolan2005automatically}, QNLI \cite{rajpurkar2016squad}, QQP \cite{chen2018quora}, and MNLI \cite{williams2018broad} in GLUE benchmark and IMDB reviews \cite{maas2011learning}. We follow the standard evaluation protocols, fine-tune the pre-trained model on the training set, report the results on the validation set, and compare them to our baseline BERT-base.

\smallskip
\noindent \textbf{Implementation details}. We fine-tune our pre-trained model on GLUE benchmark datasets and IMDB reviews respectively and report its final performance. For larger datasets (SST-2, QNLI, QQP, MMNL, IMDB reviews), we use a batch size of 32 and the AdamW optimizer with learning rate 3e-5 and fine-tune our models for 4 epochs. For MRPC, due to the sensitivity of a smaller dataset, we follow \cite{devlin2019bert} by performing a hyperparameter search with candidate batch size $[$8, 16, 32$]$ and learning rate $[$2e-5, 3e-5, 4e-5, 5e-5$]$, and select the best validation result. As these downstream tasks do not exceed the maximum input sequence length 512, we fine-tune our model trained on an input sequence length of 512.

\smallskip
\noindent \textbf{Results}. Table~\ref{tab:devset} presents our experimental results on natural language understanding benchmarks with different tasks. Our results compares favorably to BERT-base across all downstream tasks. Further, we also experiment with fine-tuning our model using longer sequences ($n=1024$), yet the results remain almost identical to $n=512$, e.g. 93.0 vs.\ 93.2 accuracy on IMDB reviews. These results suggest that our model is able to scale linearly with input length. Additional details on longer sequences is in the supplement. 

\subsection{Long Range Arena (LRA) Benchmark}
Our last experiment evaluates our model on tasks with longer sequence lengths. We follow the LRA benchmark \cite{tay2020long} and compare our method against other efficient self-attention variants. 

\smallskip
\noindent \textbf{Datasets and metrics}. We consider the LRA benchmark \cite{tay2020long} with tasks of Listops \cite{nangia2018listops}, byte-level IMDb reviews text classification \cite{maas2011learning}, byte-level document retrieval \cite{radev2013acl}, image classification on sequences of pixels  \cite{krizhevsky2009learning}, and Pathfinder \cite{linsley2018learning}. We follow the evaluation protocol from \cite{tay2020long}, including the train/test splits, and report the classification accuracy for each task, as well as the average accuracy across all tasks.

\smallskip
\noindent \textbf{Baselines}. We compare different self-attention methods using a same Transformer model. Our baselines consist of the vanilla self-attention \cite{vaswani2017attention}, and several recent efficient self-attention variants, including Reformer \cite{kitaev2019reformer}, Linformer \cite{wang2020linformer}, and Performer \cite{choromanski2020rethinking}. 

\smallskip
\noindent \textbf{Implementation details}. The official LRA benchmark \cite{tay2020long} is implemented in Jax/Flax \cite{frostig2018compiling}. To achieve a fair comparison to our baselines implemented in PyTorch, we reimplemented the benchmark in PyTorch and verified the results. All our experiments, including our method and all baselines, use a Transformer model with 2 layers, 64 embedding dimension, 128 hidden dimension, 2 attention heads. Mean pooling is used for all tasks. The number of hashes for Reformer is 2, the projection dimension for Linformer is 256, and random feature dimension for Performer is 256.

\smallskip
\noindent \textbf{Results}. Table~\ref{tab:lraset} compares our method to baselines. Our results are on par with the vanilla self-attention for all tasks, with comparable average accuracy (+0.18\%) but are more efficient (see Table~\ref{tab:timememory}). Importantly, our method outperforms other efficient self-attention methods, with +3.91\%, +3.36\%, +5.32\% in average accuracy against Reformer, Linformer, and Performer, respectively. We find that the model behaves favorably relative to the concurrent work of Performer across all tasks, and in general, provides a good approximation to self-attention for longer sequences.

%% file: conclusion.tex
Scaling Transformer based
models to longer sequences is desirable in both NLP as well as computer vision, and it will involve identifying 
ways to mitigate its compute and memory requirements. Within the last year, this need has led to a number of results describing how randomized numerical linear algebra schemes based on random projections and low rank assumptions can help \cite{katharopoulos-et-al-2020,wang2020linformer,Beltagy2020Longformer,zaheer2020big}. Here, we approach this task differently by showing how the Nystr\"{o}m method, a widely used 
strategy for matrix approximation, can be adapted 
and deployed within a deep Transformer architecture to provide an efficient approximation of self attention. We show that our design choices and modifications enable all key operations to be mapped to popular deep learning libraries conveniently. 
The algorithm maintains the performance profile of other self-attention approximations in the literature but offers additional benefit of resource utilization, and is  
a step towards building Transformer models on very long sequences. Our code/supp is available at \textit{\url{https://github.com/mlpen/Nystromformer}}. 